\documentclass[journal=jcim,manuscript=article]{achemso}

\usepackage[version=3]{mhchem} 

\usepackage{array}
\usepackage{amsmath}
\usepackage{amsfonts}
\usepackage{graphicx}
\usepackage{bm}
\usepackage{bbm}
\usepackage{booktabs}
\usepackage{xcolor} 
\usepackage{booktabs, tabularx, threeparttable} 
\usepackage{xspace}
\usepackage{hyperref}
\usepackage{url}
\usepackage{cleveref}
\usepackage{dsfont}
\usepackage{textalpha}
\usepackage{caption}
\usepackage{comment}
\usepackage{booktabs}

\author{Vincent Fan}
\affiliation{Computer Science and Artificial Intelligence Laboratory, Massachusetts Institute of Technology, Cambridge, MA, 02139}
\email{vincentf@mit.edu}
\author{Regina Barzilay}
\email{regina@csail.mit.edu}
\affiliation{Computer Science and Artificial Intelligence Laboratory, Massachusetts Institute of Technology, Cambridge, MA, 02139}
\alsoaffiliation{Department of Electrical Engineering and Computer Science, Massachusetts Institute of Technology, Cambridge, MA, 02139}

\title{AssayMatch: Learning to Select Data for Molecular Activity Models}

\newcommand{\ours}{\color{black}{\text{AssayMatch}}\color{black}\xspace}

\abbreviations{IR,NMR,UV}

\begin{document}


\begin{abstract}
The performance of machine learning models in drug discovery is highly dependent on the quality and consistency of the underlying training data. Due to limitations in dataset sizes, many models are trained by aggregating bioactivity data from diverse sources, including public databases such as ChEMBL. However, this approach often introduces significant noise due to variability in experimental protocols. We introduce \ours, a framework for data selection that builds smaller, more homogenous training sets attuned to the test set of interest. \ours leverages data attribution methods to quantify the contribution of each training assay to model performance. These attribution scores are used to finetune language embeddings of text-based assay descriptions to capture not just semantic similarity, but also the compatibility between assays. Unlike existing data attribution methods, our approach enables data selection for a test set with unknown labels, mirroring real-world drug discovery campaigns where the activities of candidate molecules are not known in advance. At test time, embeddings finetuned with \ours are used to rank all available training data. We demonstrate that models trained on data selected by \ours are able to surpass the performance of the model trained on the complete dataset, highlighting its ability to effectively filter out harmful or noisy experiments. We perform experiments on two common machine learning architectures and see increased prediction capability over a strong language-only baseline for 9/12 model-target pairs. \ours provides a data-driven mechanism to curate higher-quality datasets, reducing noise from incompatible experiments and improving the predictive power and data efficiency of models for drug discovery. \ours is available at \href{https://github.com/Ozymandias314/AssayMatch}{https://github.com/Ozymandias314/AssayMatch}.
\end{abstract}

\section{Introduction}

Property prediction models are widely used for virtual screening in modern drug discovery. However, model performance can vary substantially depending on the target or endpoint being modeled. A major factor influencing prediction accuracy is the quality of the training data, which consists of molecules paired with a property value. When large amounts of diverse and consistently measured experimental data are available, the models learn generalizable and robust features. For many properties, such comprehensive data is not readily accessible and researchers commonly combine smaller datasets instead. These measurements are recorded by laboratories or manually curated from literature and deposited in publicly available resources such as ChEMBL \cite{gaultonChEMBLLargescaleBioactivity2012a}. While aggregated datasets readily increase the size of training data, they present significant challenges due to well-documented inconsistencies between assays.

\begin{figure}[t]
    \centering
    \includegraphics[width=\linewidth]{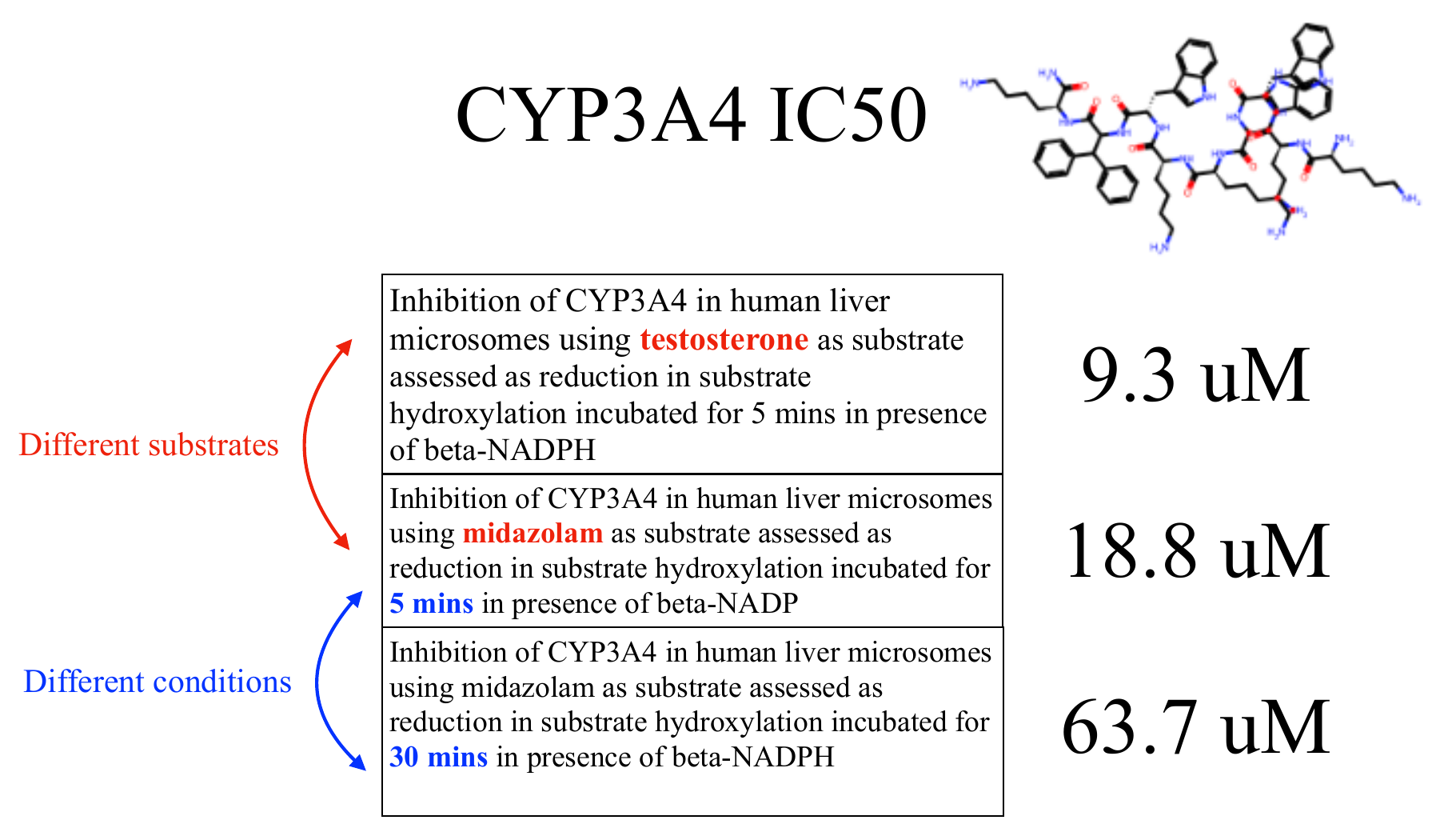}
    \caption{CYP3A4 IC50 measurements for the same molecule taken from ChEMBL assays 2296579, 2296580, 2296590. The measurements differ vastly depending on specific details in the assay description, such as the probe substrate or timing conditions.}
    \label{fig:duplicate}
\end{figure}

Combining data from heterogeneous experimental settings introduces confounding factors that obscure patterns and hinder generalization. Landrum et al. \cite{landrumCombiningIC50Ki2024a} analyzed IC50 affinity data in ChEMBL and demonstrated that over 25\% of reported measurements for the same target and small molecule differ by more than one order of magnitude. Figure \ref{fig:duplicate} depicts three inconsistent IC50 measurements of the same molecule with respect to Cytochrome P450 3A4 along with their assay descriptions in ChEMBL. There are over 1000 unique assay descriptions recorded for this target, reflecting substantial variability in experimental protocols, including differences in timing, substrate selection, expression systems, and detection methods.

Past research has shown that incorporating extra text-based assay metadata can decrease noise or improve model performance. Niu et al. extract keywords from assay descriptions in ChEMBL for manual review by experts\cite{niuPharmaBenchEnhancingADMET2024}. They generated a filtered set of data exhibiting reduced inter-assay variation. Other approaches automatically combine and filter assays for training through heuristics based on measurements and metadata. However, these rule-based consistency checks are often challenging to design due to sparse overlap of identical compounds between assays. We find that fewer than 0.5\% of ChEMBL assay pairs share even a single molecule. In the ``maximal curation" scheme devised by Landrum et al., over 99\% of all IC50 data is discarded to pass hand crafted rules. Schoenmaker et al. offer a deep learning approach by directly including language embeddings of assay context as input to their bioactivity model, showing improvement for specific targets\cite{schoenmakerAssayAwareBioactivityModelers2025}. In contrast, CLAMP, TwinBooster, and MoleculeSTM use machine-learning models that leverage unsupervised relationships within text descriptions to enhance molecular representations \cite{seidlEnhancingActivityPrediction2023, schuhTwinBoosterSynergisingLarge2024, liuMultimodalMoleculeStructuretext2024}. However, language models are not explicitly trained to reflect the differences between chemical assays. The descriptions in Figure \ref{fig:duplicate} are worded almost identically and have very similar language model embeddings, but clearly produce disparate readouts. 
Another orthogonal assay-aware approach is to treat each batch as a completely separate entity, and only learn to model and predict activity differences of compounds within the same assay \cite{passaroBoltz2AccurateEfficient2025,fengFoundationModelBioactivity2023}.

In this work we present \ours, a model agnostic framework for data selection designed to learn the functional relationships between different text-based assay conditions using data attribution. \ours takes an unseen assay description as input and generates a ranking of all training assays based on their predicted contribution to model performance on the test assay. This contrasts with traditional data attribution methods, which are unsuitable for data selection, since they require the true label of every point in advance. However, signals from data attribution methods are able to explain the effects of different assays on model performance.  \ours utilizes these signals to finetune the language embeddings of a large curated set of ChEMBL metadata without inspecting the held-out test set to improve upon simple semantic similarity. \ours enables the construction of smaller, more efficient, and higher-quality training sets for improved model performance and generalizability. Using \ours to select data for two common machine learning architectures, we were able to surpass the performance models trained on the full dataset by removing the 10\% of samples identified as least informative among other gains in data efficiency. We make our implementation available at \href{https://github.com/Ozymandias314/AssayMatch}{https://github.com/Ozymandias314/AssayMatch}.

\section{Related Work}



\textbf{Data Attribution Methods} Our work relates to research in data attribution. These methods seek to quantify the contribution of each point in the training set to the model's performance. They can identify points that introduce noise to the training process. Specifically, they build upon classical influence functions, which measure the leave-one-out effect of removing each individual training point from the dataset. In deep learning, however, training a separate model with each individual point held out from the dataset is computationally prohibitive. Therefore, most of the existing methods focus on estimating quantities in the mathematical formula for leave-one-out scores. Koh et al. first extended influence functions to simple deep learning models by making first and second order approximations on the loss function to estimate the leave-one-out effect\cite{kohUnderstandingBlackboxPredictions2020}. Other approaches repeatedly sample different subsets of the training set to empirically estimate the leave-one-out effect. Sampling based approaches achieve better attribution performance for more complex machine learning architectures, but require high computational cost \cite{ghorbaniDataShapleyEquitable2019, feldmanWhatNeuralNetworks2020, ilyasDatamodelsPredictingPredictions2022a}. TRAK is a recent method that refines both approaches by utilizing more nuanced approximations of the loss function to reduce the number of subsamples, yielding an efficient data attribution method with strong performance for large models\cite{parkTRAKAttributingModel2023}. 

However, current data attribution methods are unapplicable for our task of prospective data selection, as we do not know the labels of the test set in advance. Existing methods are developed for interpretability, as they are meant to explain whether or not a specific training example was useful towards making a prediction. Therefore, to analyze the prediction for a given test point, these methods utilize both its input features and its ground-truth label. Some existing selection strategies using data attribution will retrain the model after removing the most harmful points with respect to a validation dataset or a test set with known labels \cite{wangDataShapleyOne2025}. These approaches implicitly require that the validation and test set come from the same data distribution. In our problem setting, the features of the test assay are completely unseen and assumed to be from distinct experimental protocols, which necessitates a fundamentally different approach to data selection that operates without access to test-time labels. Thus, we introduce a framework where we are able to transfer the relationships revealed by TRAK on a set of training assays into the space of natural language assay descriptions. By finetuning assay embeddings to reflect TRAK-derived compatibility, \ours enables assay–assay relationships learned from labeled training data to generalize to unseen assays where only textual metadata is available. This makes it possible to rank candidate training data for a new assay without requiring access to its experimental measurements.

 \textbf{Language Embeddings} Language embeddings, which represent text as dense vectors \cite{mikolovEfficientEstimationWord2013, devlinBERTPretrainingDeep2019, leeGeminiEmbeddingGeneralizable2025}, are used as additional guidance for our task. A critical property of these embeddings is that their vector geometry captures semantic relationships, meaning that similar semantic concepts are close to each other in embedding space. Many previous methods enhance the structure of available off the shelf embeddings through contrastive learning or finetuning with another modality \cite{qianPredictiveChemistryAugmented2023, liuMultimodalMoleculeStructuretext2024, edwardsText2MolCrossModalMolecule2021}. For example, MoleculeSTM aligns traditional encodings of molecules with language embeddings of associated text descriptions to produce improved molecular representations for property prediction \cite{liuMultimodalMoleculeStructuretext2024}. Qian et al. utilize curated pairs of chemical reactions and conditions to align their respective embeddings for the purpose of reaction condition recommendation \cite{qianPredictiveChemistryAugmented2023}. These methods are meant to take into account relationships present in the embeddings of both modalities, thus increasing the overall expressivity of the new representation. However, neither molecular embeddings nor natural language embeddings are meant to explicitly reflect the training value of specific examples. In \ours, we utilize statistics computed from data attribution methods as a more explicit signal to finetune language embeddings instead of relying on unsupervised relationships between pairs of modalities. We use frontier large language model embeddings as a starting point, since they capture broad semantic patterns. This approach ensures that the learned representations are optimized for the downstream data selection task.

\section{Method Description}

Given a training set containing assays with descriptions $\{A_1, \dots, A_n\}$ and a new test assay with experimental description $A_\text{test}$, \ours identifies a training subset which is most informative. We select the subset by computing $\text{AssayMatchScore}(A_i, A_\text{test})$ for each training assay and creating a ranking, where a higher score indicates a more favorable assay for machine learning model training. Currently, the method is agnostic to the labels and the molecular contents of assays held out for testing, and only depends on learned relationships between different assay conditions. When we select a training assay, we include all of its measurements. \ours is compatible with arbitrary deep learning architectures, as long as TRAK scores can be computed.

 \begin{figure}[!h]
    \centering
    \includegraphics[width=\linewidth]{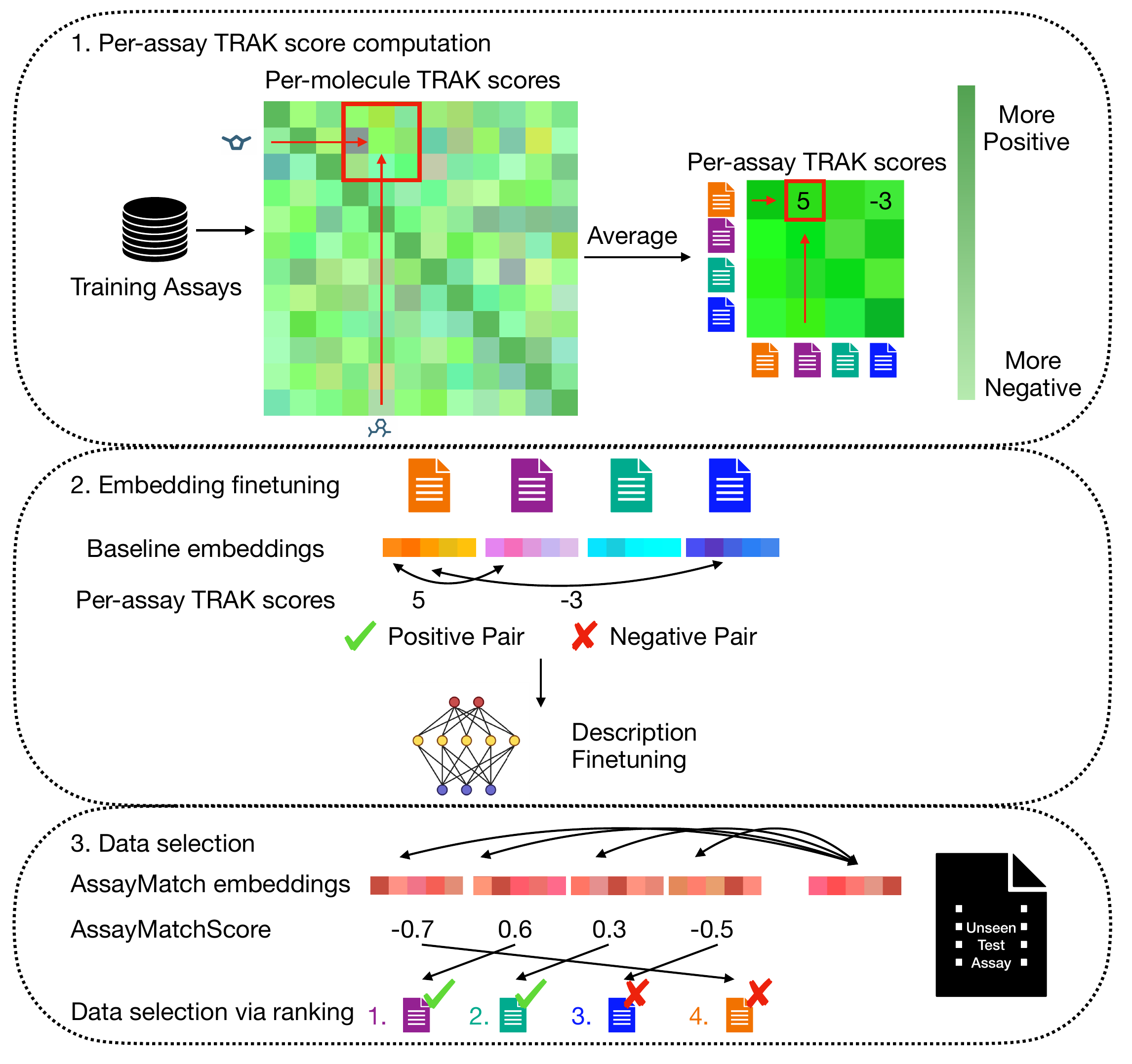}
    \caption{Overview of \ours. First, per assay TRAK scores are computed to elucidate assay to assay relationships. Next, language embeddings of experimental descriptions are finetuned to reflect these relationships. Lastly, the finetuned embeddings are used to rank data with respect to unseen assay descriptions at test time.}
    \label{fig:overview}
\end{figure}

\ours employs a three stage process where we compute data attribution scores, finetune the language embeddings of training assay descriptions, and then perform data selection as depicted in Figure \ref{fig:overview}. First, we run TRAK over all training assays, and aggregate the scores to the assay level. Next, to finetune the baseline language embeddings, we sample a large number of positive and negative assay pairs as defined by the per-assay TRAK scores. The finetuning process pushes positive pairs of descriptions to be closer in the embedding space. At selection time, we compute the finetuned embedding for an unseen test assay description and compute $\text{AssayMatchScore}$ to create a ranking across all training assays. We greedily select training assays in the order of this ranking to create a training set of desired size.

\subsection{Per-assay TRAK score computation}

We compute per-assay TRAK scores to understand the relationship between different assays for the same target. First, for a given target, we combine all of its training assays and implement TRAK to obtain attribution scores. A detailed description of the implementations are deferred to the supporting information. $TRAK(m_1, m_2)$ quantifies the effect of training on a molecule $m_1$ and evaluating on a molecule $m_2$. A larger score corresponds to a more positive contribution. Next, given a pair of assays $A_1$ containing molecules $\{m_{1,1}, \dots, m_{1, n}\}$ and $A_2$ containing molecules $\{m_{2,1},\dots, m_{2,k}\}$, we define the per-assay TRAK score as the arithmetic mean of all pairwise per molecule TRAK scores as in equation \ref{eq:perassay}. \begin{equation} \label{eq:perassay}
    \text{TRAK}_\text{Assay}(A_1, A_2) = \frac{1}{|A_1|\cdot|A_2|}\sum_{i=0,j=0}^{i=n, j = k}\text{TRAK}(m_{1,i}, m_{2,j})
\end{equation} This captures the aggregate effect of training on data from $A_1$ and evaluating on the measurements for molecules in $A_2$. If the score defined by $\text{TRAK}_\text{Assay}$ is positive, then the effect is beneficial and vice versa.

\subsection{Embedding Finetuning}

 The baseline language embeddings of each assay description are further enhanced by finetuning with a contrastive learning signal derived through TRAK. We create triples of embeddings for training, consisting of an anchor embedding $e_A$, a positive embedding $e_P$, and a negative embedding $e_N$. We seek to decrease the distance between $e_A$ and $e_P$ while increasing the distance between $e_A$ and $e_N$. For an anchor assay $e_A$, we order all remaining assays by $\text{TRAK}_\text{Assay}(\cdot, A)$, such that a higher rank indicates a more beneficial assay for training. Assays in the first half of the ranking are considered to be positive and assays in the second half are negative. This ensures that $e_P$ always corresponds to a more useful assay than $e_N$ as it appears in the ranking first. The model is finetuned with triplet margin loss \begin{equation}
    L = \max(0, d(f(e_A), f(e_P)) - d(f(e_A),f(e_N)) + m)
\end{equation}
where $d$ is a distance function and $m$ is the margin. The loss encourages the difference between the distance of the anchor to the positive and negative examples to be at least the margin $m$, effectively pushing assays with high TRAK scores to be closer to the anchor. 

\subsection{Data selection}

Given a test assay $A_\text{test}$, we use its finetuned embedding $f(e_{A_\text{test}})$ to select data from the set of available training assays $\{A_1,\dots,A_n\}$. Specifically, the \ours similarity score is given by the dot product of finetuned description embeddings \begin{equation}
    \text{AssayMatchScore}(A_i, A_\text{test}) = f(e_{A_i})\cdot f(e_{A_\text{test}})
\end{equation} In our experiments, we rank all training assays using $\text{AssayMatchScore}$ and select the highest scoring assays until we have reached a desired training set size.

\section{Experiment}

\subsection{Implementation Details}

For an assay $A$, we compute the baseline embedding of its ChEMBL description $e_A$ using the \texttt{text-embedding-004} model from Gemini\cite{leeGeminiEmbeddingGeneralizable2025} with dimension 768. To finetune the embedding, we select $f$ to be a 2-layer neural network with ReLU activations. The model is trained for 10 epochs on one NVIDIA A6000 GPU with learning rate \texttt{1e-4}, batch size 512, and margin $d$ set to 0.1. We measure the binary classification performance on selected datasets with Chemprop and SMILES Transformer \cite{yangAnalyzingLearnedMolecular2019b, hondaSMILESTransformerPretrained2019}. Chemprop is a widely used graph neural network for small molecule property prediction with molecular graph structures as inputs. SMILES Transformer instead directly operates on a tokenized representation of the molecule. Together, these architectures represent the most common deep learning approaches for molecular modeling. Since there is no validation set for our task, we initialize both models with the default hyperparameters suggested by the original authors. 

\subsection{Evaluation Setting}

To evaluate \ours, we process a large subset of ChEMBL IC50 data and choose six diverse targets with the largest number of measurements or assay descriptions identified by text equality. We group all data points with the same experimental description together and split the data into training and test sets with a 90:10 ratio. This split ensures that no descriptions in the test set are leaked in the training set. We measure the binary classification performance of selected models with AUROC, where a molecule is considered active if it has an IC50 value less than 1$\mu$M. Additional details for the data deduplication and standardization process are provided in the supporting information.
\begin{figure}[!h]
    \centering
    \includegraphics[width=0.9\linewidth]{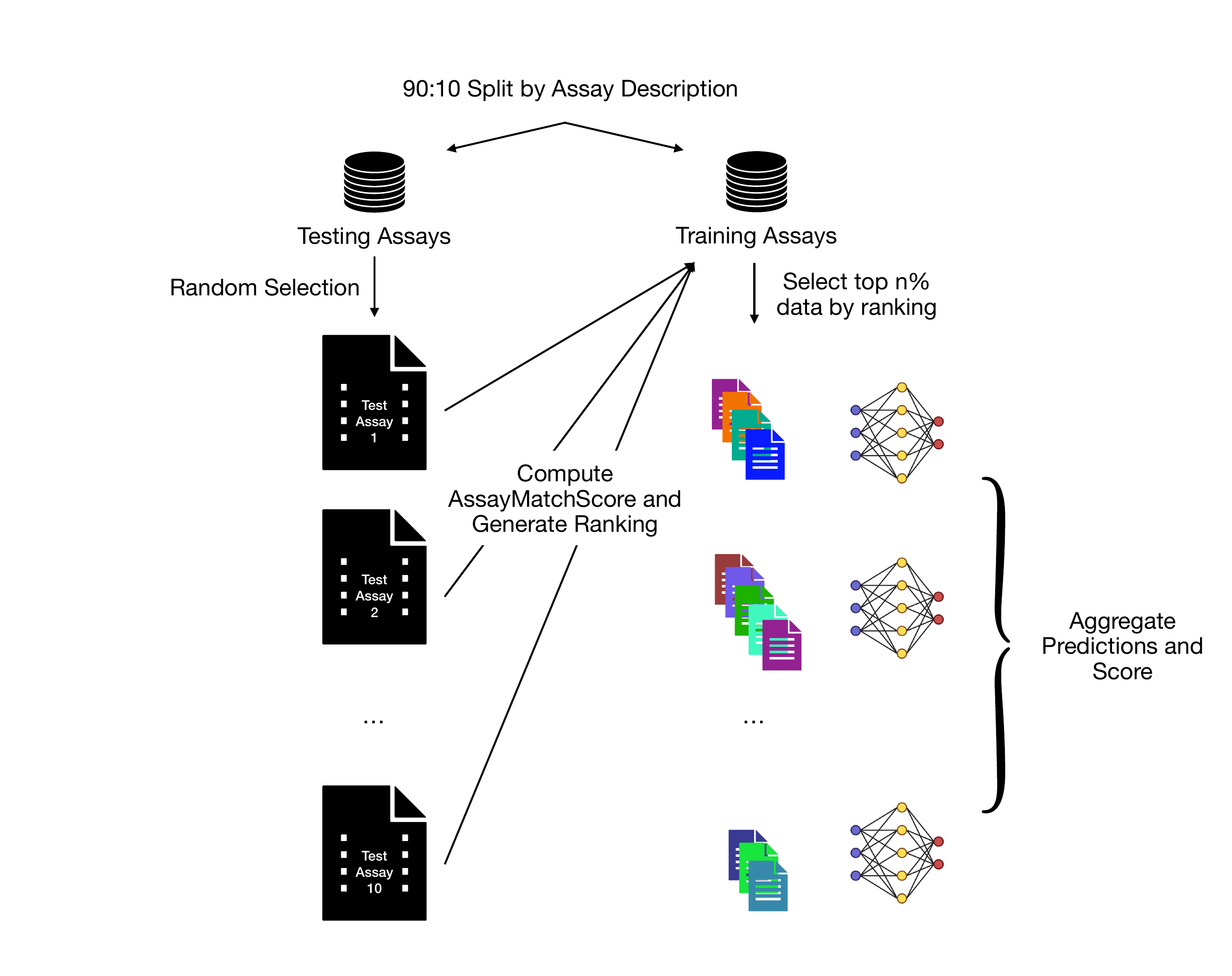}
    \caption{To evaluate \ours, we randomly select 10 assay descriptions for each target. \ours selects a separate dataset for each test assay, which we train a separate model on. The predictions are aggregated and scored by AUROC to construct a learning curve.}
    \label{fig:evaluation}
\end{figure}

The evaluation procedure is illustrated in Figure \ref{fig:evaluation}. For each target, we randomly select ten assay descriptions from the set of testing assays. For every test assay, \ours ranks and selects subsets of training data that increase in size by 10\% increments of the full training set size. A separate model is trained on each subset using default classification parameters. We compute the micro-averaged AUROC of the predictions from each model. Since \ours selects data based on the descriptions of individual test assays, training outcomes can vary due to small test assay sizes or extremely mismatched molecular compositions. Thus, we repeat this procedure with 15 independent train-test splits performing five runs per split and report the averaged metrics across all six targets. The performance of \ours is summarized with a learning curve, which compares the AUROC against the fraction of training data used. Learning curves are commonly used to capture how predictive accuracy improves as more training data points are included under various strategies\cite{vieringShapeLearningCurves2022}. We calculate the Area Under the Learning Curve (AULC) to provide a direct quantitative measure of the data efficiency achieved by \ours.  

We further benchmark \ours against several alternative dataset selection strategies. \begin{itemize}
    \item \textbf{Random}: Training assays are chosen uniformly at random, according to the same dataset size thresholds which increase uniformly by 10\%. 
    \item \textbf{Baseline embedding}: Training assays are ranked by the dot product similarity of their pre-finetuned embeddings, and then selected to meet the same dataset size thresholds. We construct a learning curve as well. 
    \item \textbf{BioAssay Ontology}: A single training set is selected based on exact match of the BioAssay Ontology \cite{abeyruwanEvolvingBioAssayOntology2014} annotation. The BioAssay Ontology provides a controlled vocabulary for classifying assays by features such as design, detection technology, and substrate. This baseline serves as a human-curated counterpart to embedding-based selection, relying on expert-defined metadata rather than learned representations. We consider the BAO-based approach particularly suitable because it reflects a structured, interpretable classification of assays while still ensuring a sufficiently large training set—unlike other metadata-based filters that can be overly restrictive and yield too few data points. Since the BAO selection produces a single fixed-size training set per test assay, we do not construct a learning curve for this baseline.
\end{itemize}

\subsection{Results}

\begin{figure}[!h]
    \centering
    \includegraphics[width=0.9\linewidth]{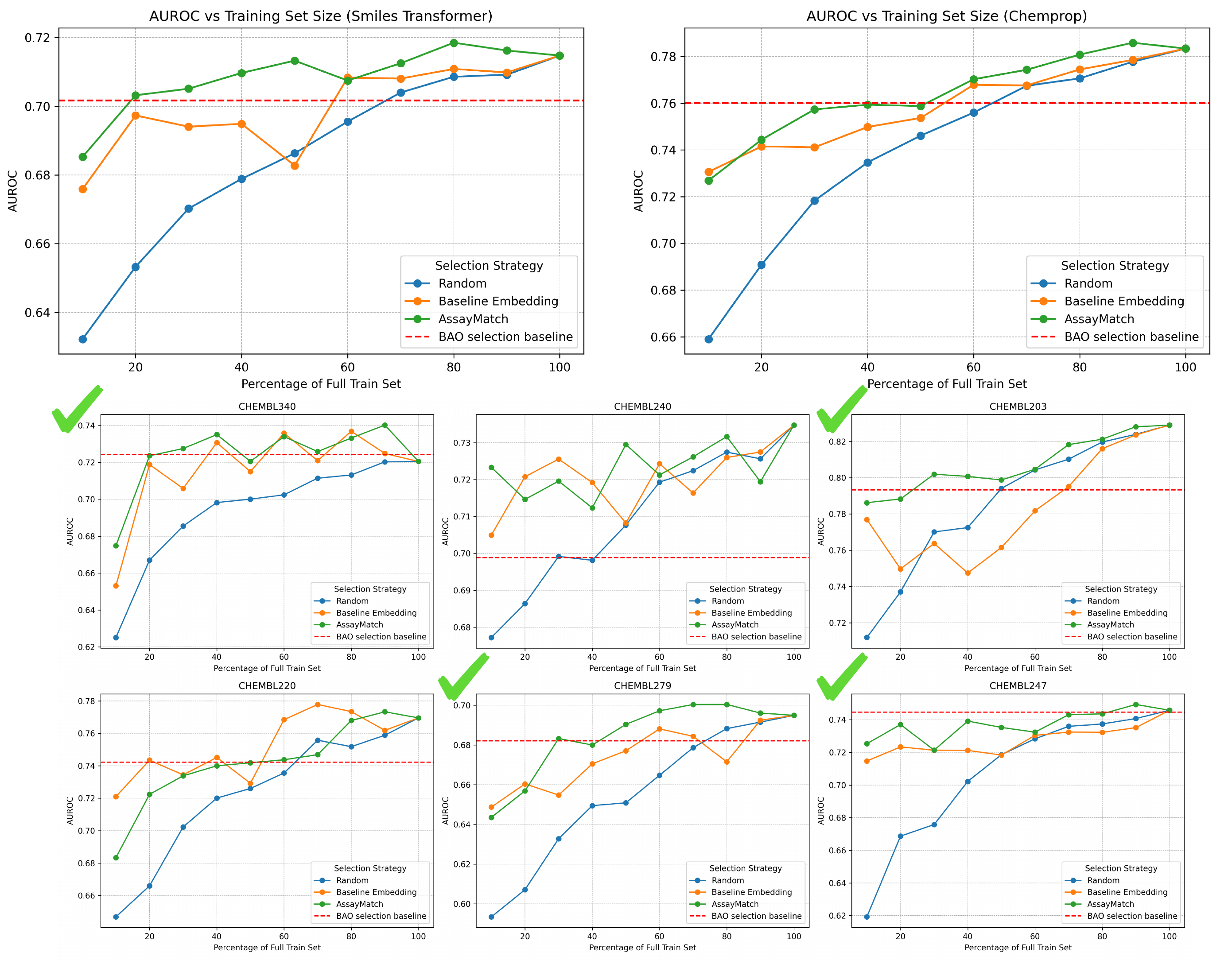}
    \caption{Microaveraged AUC scores of Chemprop and SMILES Transformer trained on subsets of different size as selected according to a random baseline, the original language embeddings, and \ours embeddings. The performance of the model trained on the full available dataset is represented when size = 100\%. The performance obtained by selecting all training assays that share the same BioAssay Ontology (BAO) label as the test assay is represented by the red dashed line. Results for each individual target (averaged over both architectures) displayed below. \ours is the best strategy for 4/6 targets.}
    \label{fig:train_set}
\end{figure}

\begin{table}[t]
    \centering
    \resizebox{0.99\linewidth}{!}{
\begin{tabular}{llllllll}
\toprule
AULC & &Chemprop&& && SMILES Transformer&\\
\midrule

& \ours & Embedding & Random && \ours & Embedding & Random \\
\cline{2-4}\cline{6-8}    \\
Overall & \textbf{74.63 (p = 0.007)} & 74.12 & 72.02 && \textbf{69.59 (p = $\mathbf{1\times 10^{-7}}$)} & 68.64 & 67.03 \\
\midrule
\textit{Evaluation of individual targets} \\
CHEMBL340 & 74.04 & 74.03 & 71.79 && \textbf{68.03 (p = 0.001)} & 66.73 & 64.06 \\
CHEMBL240 &  72.76 & 73.32 & 72.62 && \textbf{69.23 (p = 0.009)} & 68.03 & 66.24 \\
CHEMBL203 & \textbf{81.18 (p = $\mathbf{7\times 10^{-8}}$)} & 77.83 & 78.59 && \textbf{76.35 (p = $\mathbf{9\times 10^{-5}}$)} & 74.57 & 74.34 \\
CHEMBL220 & 73.82 & \textbf{75.68 (p = $\mathbf{1\times 10^{-6}}$)} & 71.71 && 71.00 & 71.54 & 69.02 \\
CHEMBL247 & \textbf{74.92 (p = 0.005)} & 73.45 & 70.44 && 69.51 & 68.95 & 67.48 \\
CHEMBL279 &  71.09 & 70.40 & 66.94 && \textbf{63.43 (p = 0.003)} & 61.99 & 61.09 \\
\bottomrule
\end{tabular}
}
    \caption{AULC scores for different dataset selection strategies with respect to Chemprop and SMILES Transformer, with statistically significant comparisons in \textbf{bold}. \ours demonstrates clear improvement over the embedding baseline for both architectures in the overall average and produces the higher AULC for 9/12 model-target pairs, including 6/7 significant comparisons.}
    \label{tab:aulc}
\end{table}

\ours is able to select training subsets with several advantages. First, \ours achieves AUROCs of 78.58 and 71.85 for Chemprop and SMILES Transformer respectively, which both surpass the performance of models trained on the entire training dataset. On the other hand, selection via the baseline embeddings is slightly worse than \ours and on average only matches the random baseline at larger dataset sizes. This suggests that \ours is able to rank the most harmful data more accurately, and that removing the bottom 10\%-20\% data is beneficial despite the reduction in dataset size. Additionally, we observe significant p-values when performing a paired t-test on AUROCs from \ours datasets and baseline embedding datasets ($5\times 10^{-5}$ and $9\times 10^{-11}$ for Chemprop and SMILES Transformer respectively). In the low data regime, our method substantially outperforms random selection. As seen in the learning curves displayed in Figure \ref{fig:train_set}, when the training set is restricted to 10\% or 20\% of the entire pool of data, models trained on datasets selected with \ours embeddings outperform their random counterparts by more than 5 AUROC. This further reflects the fact that language based embeddings of assay descriptions are correlated with data similarity. \ours also exhibits the best data efficiency properties. On average, models trained on \ours datasets approach the performance of training on the entire dataset around a size threshold of 50\%-70\%. Another quantitative measurement of data efficiency is the Area Under the Learning Curve, which we calculate for both models in Table \ref{tab:aulc}. \ours has the best overall AULC for both models, and also the highest AULC for 9 out of 12 target model-pairings, showing that the model is robust across targets. These results underscore the efficiency of our method, as identical or even superior performance can be achieved with fewer training assays, reducing noise from potentially incompatible experiments.

As \ours produces a continuous ranking of every training assay, we can perform proportion-based subset selection rather than relying on the strict grouping imposed by ontology labels. Selecting by BioAssay Ontology (BAO) relies on human-curated annotations, which provide a strong, systematic signal of assay similarity but requires annotations to match exactly. The average size of a BAO-selected training set is 51.1\% of the full dataset, and \ours recapitulates this curated signal, matching BAO-level performance in the 30\%-40\% size range while also permitting fine-grained control over which and how many assays are included. This flexibility allows adjusting the inclusion cutoff to trade off between training set size and model performance.

\begin{figure}[t]
    \centering
    \includegraphics[width=\linewidth]{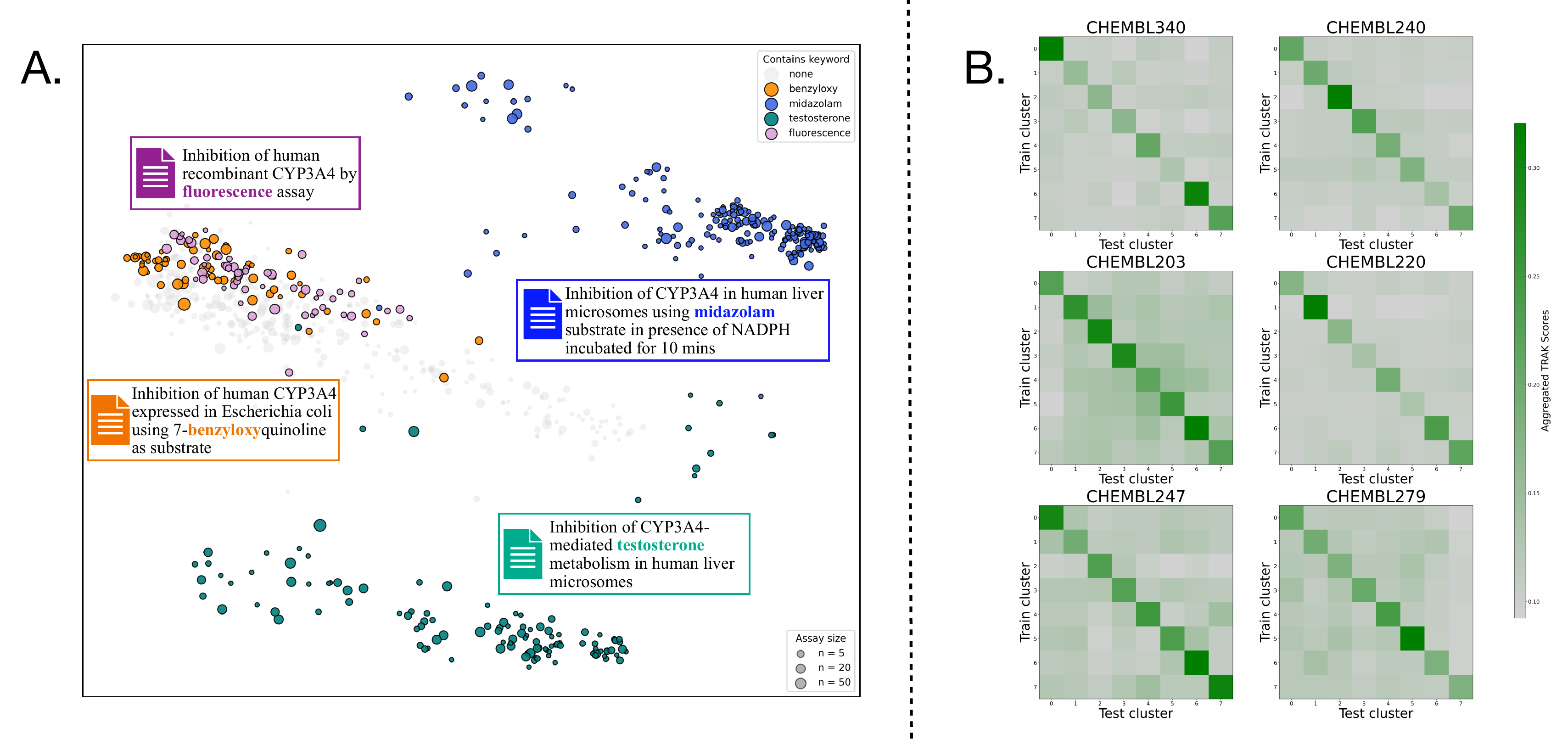}
    \caption{ A: PCA of assay description embeddings with \texttt{text-embedding-004} model on all ChEMBL assay descriptions for CYP3A4. Descriptions including similar keywords are clustered together and highlighted. B: Aggregated pairwise TRAK scores between $k$-means clusters of assay embeddings. Inter-cluster TRAK scores along the main diagonal are significantly higher. }
    \label{fig:embedding}
\end{figure}

\subsection{Analysis}

We further study how the \ours framework impacts the geometry of text embeddings. First, we examine the relationships encoded in the baseline embeddings before finetuning \ours. In the PCA plot of embeddings for Cytochrome P450 3A4 in Figure \ref{fig:embedding}A, descriptions containing representative keywords are clustered close to each other. This qualitatively reaffirms that language embeddings are able to disambiguate between broad scientific terms \cite{schoenmakerAssayAwareBioactivityModelers2025}. Moreover, the baseline embeddings also show some correlation to data attribution relationships. For each of the six targets in our test set, we perform a $k$-means clustering on their description embeddings with $k=8$ and average the per-assay TRAK scores between pairs of clusters. As depicted in Figure \ref{fig:embedding}B, scores along the main diagonal are consistently higher than off-diagonal terms. The main diagonal represents TRAK scores averaged within a cluster of semantically similar descriptions. This corresponds with the intuition that measurements from assays with similar descriptions are generally more transferable and useful for predicting related measurements. Conversely, measurements from assays with dissimilar descriptions capture distinct biological or experimental contexts and are therefore less informative for each other. Interestingly, we also observe that the intra-cluster scores that are the lowest often correspond to the least informative descriptions. For example, cluster 6 for CHEMBL340 contains generic placeholder descriptions such as ``Inhibition of recombinant CYP3A4 (unknown origin)" which contain no additional details about the experimental protocol.
\begin{figure}[!h]
    \centering
    \includegraphics[width=\linewidth]{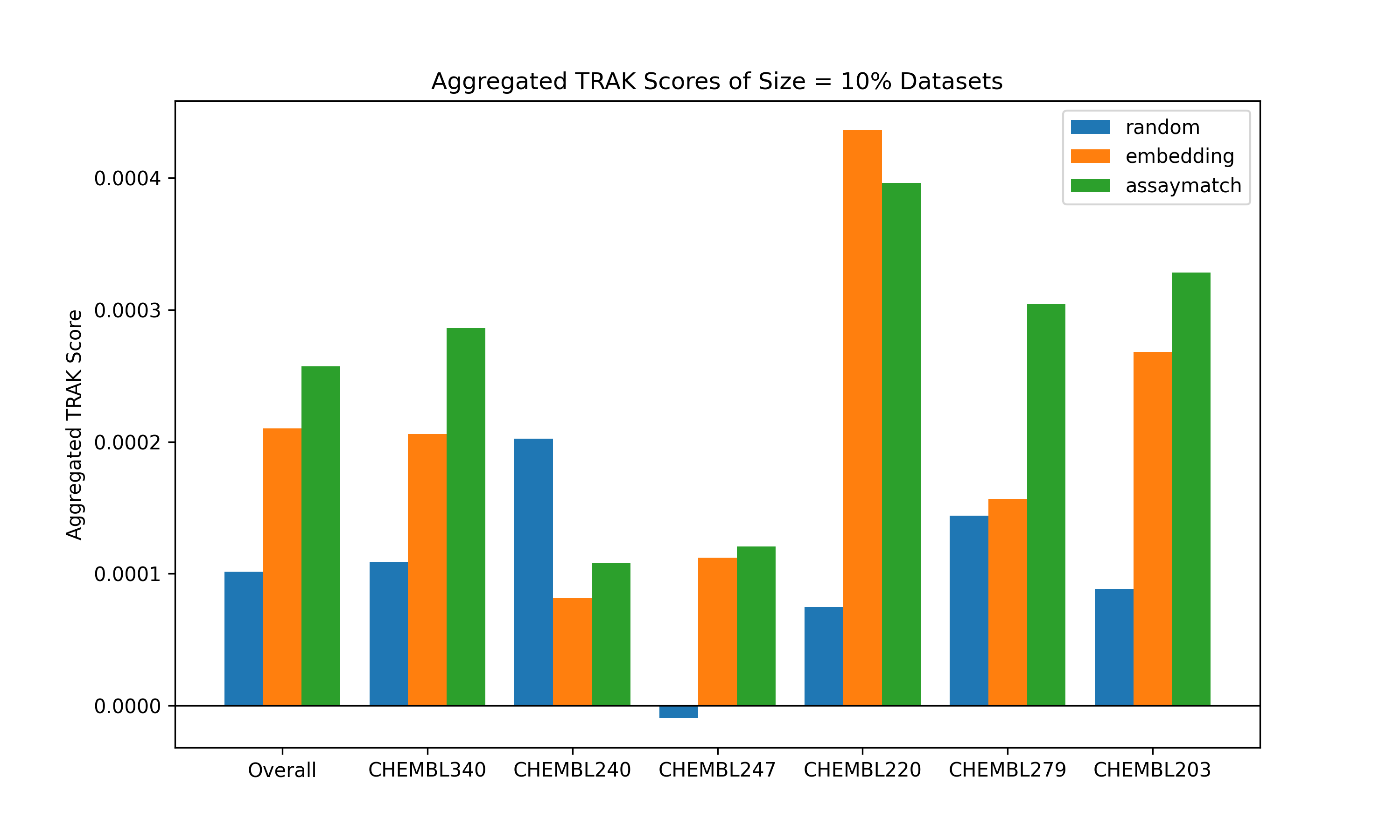}
    \caption{Average TRAK scores between test sets and training sets with size = 10\% according to different selection strategies. Embeddings finetuned by \ours cause assays with higher TRAK score to be selected more often.}
    \label{fig:trak_scores}
\end{figure}

Next, we will show that training on TRAK-derived signals concretely aligns \ours embeddings with true TRAK scores by computing the average TRAK score between every train set and test set pair produced by each strategy at the 10\% size threshold. Specifically, given a testing assay $A_\text{Test}$ and the selected training dataset consisting of assays $\{A_1,\dots, A_n\}$, we compute the average of $\text{TRAK}_{\text{Assay}}(A_i, A_\text{Test})$ weighted by size. While the absolute magnitudes of TRAK scores are difficult to interpret, their relative magnitudes are directly proportional to the performance of the model as predicted by TRAK. As seen in Figure \ref{fig:trak_scores}, the training assays selected by \ours produce the highest overall average TRAK score, suggesting that the finetuned embeddings do correspond to more compatible assays as measured by data attribution. There is 2-3x fold increase in TRAK score for \ours assays compared to randomly selected assays, which are not inherently correlated with TRAK. Moreover, we observe that the four targets on which the \ours selects assays with the highest TRAK score are the same four targets on which \ours has the best AUROC score as displayed in Figure \ref{fig:train_set}. For CHEMBL240, the baseline embeddings are the most weakly correlated with TRAK and \ours is unable to significantly improve upon them, whereas for CHEMBL220 the baseline embeddings are the most strongly correlated with TRAK and \ours produces slightly worse scores. 

\begin{figure}[!h]
    \centering
    \includegraphics[width=\linewidth]{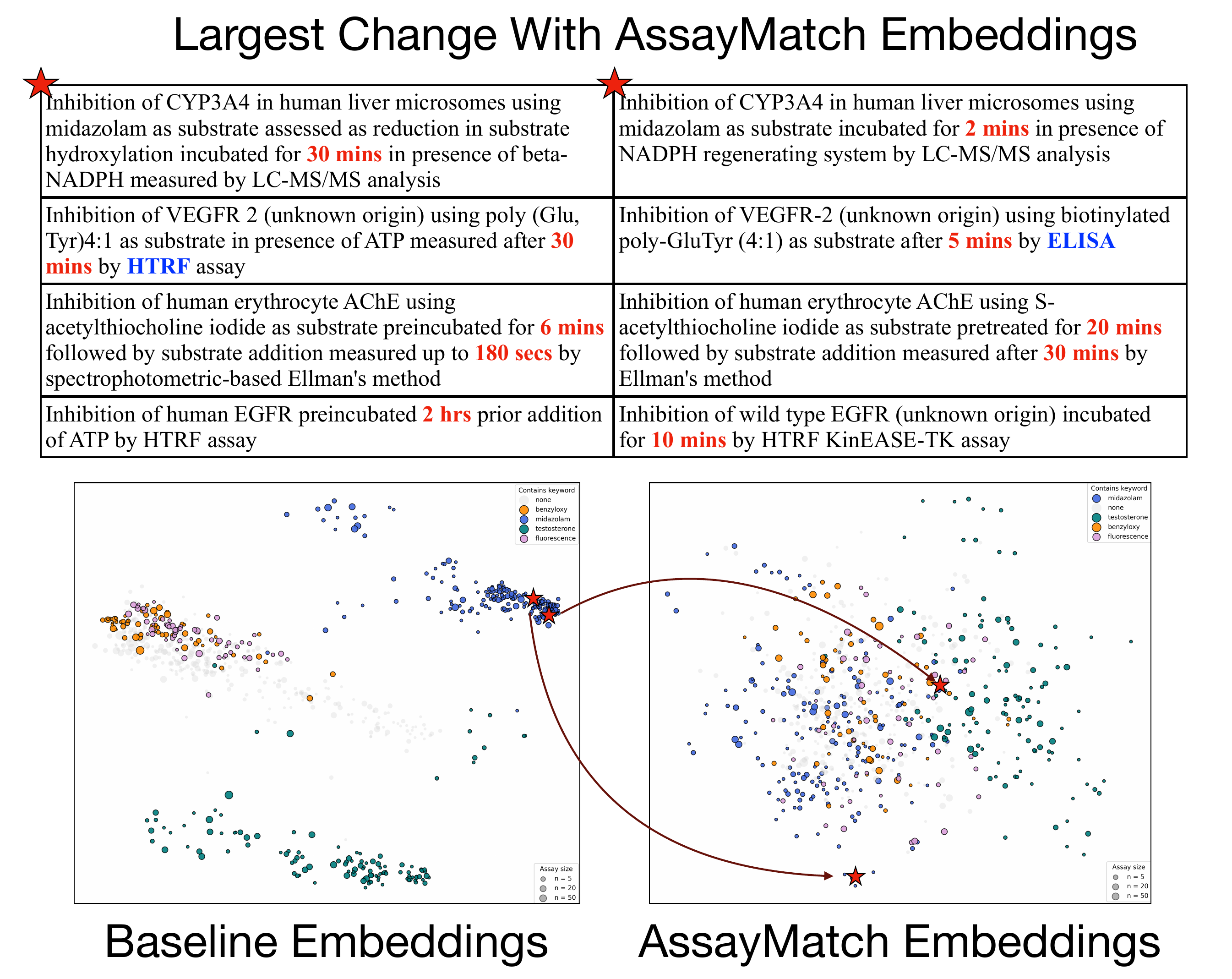}
    \caption{Pairs of assay descriptions whose embedding distances exhibited the most change after finetuning by \ours. Embeddings of CYP3A4 assay descriptions before and after finetuning, with the pair of assay descriptions labeled by red stars.}
    \label{fig:largest_change}
\end{figure}

To further probe the behavior of \ours, we qualitatively analyze the geometric changes in embeddings before and after finetuning. As seen in figure \ref{fig:largest_change}, the \ours embeddings still preserve some clustering present in the original semantic embeddings, but each individual cluster is overall more spread out. We also inspect specific pairs of embeddings which undergo the largest change after finetuning. Specifically, given two assays $A_1, A_2$, we compute the difference $d(f(e_{A_1}), f(e_{A_2})) - d(e_{A_1}, e_{A_2})$, which measures how "far" a pair of descriptions is pushed apart by \ours from the baseline language embeddings. As displayed in Figure \ref{fig:largest_change}, we consistently find pairs of assays which are semantically nearly identical, but differ in specific conditions, especially those related to timing. In the original embedding space, these pairs of descriptions would evidently be placed very closely. We hypothesize that \ours is able to regularize the original semantic space by magnifying small but mechanistically meaningful differences. For example, the length of incubation can affect the IC50 measurement depending on the exact kinetic mechanism of the inhibitor in question.

\section{Conclusion}

In this work we present \ours, a novel framework for molecular property data selection. \ours bridges the theoretical guartantees afforded by data attribution methods with the continuous embedding space provided by language models to accurately pick data based on experimental text metadata. Using \ours embeddings to rank and select training assays yields smaller, cleaner training sets and improves downstream predictive performance across multiple targets with less data. \ours yields a data-driven mechanism to reduce harmful pooling of incompatible assays and to prioritize assays that supply the most transferable signal.

\ours reflects the urgent premise to carefully filter and aggregate data as machine learning based molecular modeling becomes ubiquitous in pharmacutical workflows. To strengthen the base embeddings utilized by \ours, one could systematically mine detailed experimental protocols from the original publication associated with each assay or even patents, which are underrepresented in ChEMBL. One limitation of \ours is that we explictily ignore the molecular contents of each assay, but future improvements could incorporate this information for selection as well while preventing overfitting to a single molecular space. Furthermore, \ours currently relies on the large number of assays curated for IC50 on ChEMBL and is most applicable to situations with many individual assay descriptions. Research for low data regimes will serve as an orthogonal complement to more comprehensive curation efforts. Another limitation of \ours is that the finetuned embeddings are less interpretable than the original embeddings. Continued developments in mechanistic interpretability may provide natural language explanations for the compatibility of different assays. 

\section{Data and Software Availability}

All code and benchmarking datasets can be found at \href{https://github.com/Ozymandias314/AssayMatch}{https://github.com/Ozymandias314/AssayMatch}. Additionally, datasets used for benchmarking can be found at \href{https://doi.org/10.5281/zenodo.17656531}{https://doi.org/10.5281/zenodo.17656531}.
\begin{acknowledgement}

The authors would like to thanks Itamar Chinn, Serena Khoo, and other members of the Regina Barzilay Group for helpful discussion during the development of this work. This work was supported by the Jameel Clinic Fellowship. This work was supported by the Sanofi iDEA-TECH grant. This material is based upon work supported by the National Science Foundation Graduate Research Fellowship Program under Grant No. 2141064. Any opinions, findings, and conclusions or recommendations expressed in this material are those of the author(s) and do not necessarily reflect the views of the National Science Foundation.

\end{acknowledgement}


\bibliography{AssayTrakBib}

\end{document}